\documentclass[a4paper, 10 pt, conference]{ieeeconf}

\IEEEoverridecommandlockouts
\usepackage{adjustbox}
\usepackage{color}
\usepackage{amsfonts}
\usepackage{amsmath}
\usepackage{cite}
\usepackage{makecell}
\usepackage{siunitx}
\usepackage{gensymb}
\let\labelindent\relax
\usepackage{enumitem}
\usepackage[implicit=false]{hyperref}
\hypersetup{
    colorlinks=true,
    urlcolor=magenta,
}

\usepackage{array}
\newcolumntype{?}{!{\vrule width 2pt}}

\usepackage{booktabs}
\usepackage{multirow}

\usepackage{xcolor}

\definecolor{CommentTB}{rgb}{1.0,0,0} 
\definecolor{CommentUrgent}{rgb} {0.7,0.0,0.7} 
\definecolor{CommentHP}{rgb}{0.0,0.7,0.0} 
\definecolor{CommentLK}{rgb} {0,0.7,1.0} 

\newcommand{\pmnnotes}[1] {{\color{CommentUrgent} { PaulN: \textbf{#1}}}}
\newcommand{\todo}[1] {{\color{CommentUrgent} { TODO: \textbf{#1}}}}

\newcommand{\ignore}[1]

\renewcommand{\pmnnotes}[1]{}
\renewcommand{\todo}[1]{}
\renewcommand{\ignore}[1]{}

\DeclareMathOperator*{\argmax}{arg\,max}

\usepackage{tikz}
\usepackage{overpic}

\begin{document}

\title{Generating All the Roads to Rome: Road Layout Randomization for Improved Road Marking Segmentation}

\author{Tom Bruls, Horia Porav, Lars Kunze, and Paul Newman
\thanks{{Authors are from the Oxford Robotics Institute, Dept. Engineering Science, University of Oxford, UK. \{\texttt{tombruls}, \texttt{horia}, \texttt{lars}, \texttt{pnewman}\}\texttt{@robots.ox.ac.uk}}}
}

\maketitle

\begin{abstract}
Road markings provide guidance to traffic participants and enforce safe driving behaviour, understanding their semantic meaning is therefore paramount in (automated) driving.
However, producing the vast quantities of road marking labels required for training state-of-the-art deep networks is costly, time-consuming, and simply infeasible for every domain and condition.
In addition, training data retrieved from virtual worlds often lack the richness and complexity of the real world and consequently cannot be used directly. 
In this paper, we provide an alternative approach in which new road marking training pairs are automatically generated.
To this end, we apply principles of domain randomization to the road layout and synthesize new images from altered semantic labels. 
We demonstrate that training on these synthetic pairs improves mIoU of the segmentation of rare road marking classes during real-world deployment in complex urban environments by more than $\textbf{12}$ percentage points, while performance for other classes is retained.
This framework can easily be scaled to all domains and conditions to generate large-scale road marking datasets, while avoiding manual labelling effort.
\end{abstract}

\section{Introduction}
\label{sec:introduction}


Safety-critical systems, such as automated vehicles, need interpretable and explainable decision-making for real-world deployment.
An important aspect for improving interpretability of such systems is the ability to explain scenes semantically.
More specifically, planning the behaviour of an automated vehicle through an urban traffic environment requires understanding of the \emph{road rules}.
These are conveyed to the traffic participants by the markings painted on the road.

Although semantic reasoning about road markings is ideally performed at an object and scene level \cite{kunze2018scene}, state-of-the-art deep learning methods perform semantic segmentation at the pixel level.
This, however, requires thousands of pixel-labelled images for different environments and conditions, which is a problem for several reasons.
Firstly, it is impossible to label every pixel of every image for every city in every condition manually.
Secondly, simple data augmentation techniques \cite{cubuk2018autoaugment} (e.g. flipping, translating, adjusting contrast, etc.) do not deliver the necessary diversity to adapt to all encountered environments and conditions \cite{krajewski2019vegan}.

\begin{figure}[!t]
	\centering
	\includegraphics[width=\columnwidth]{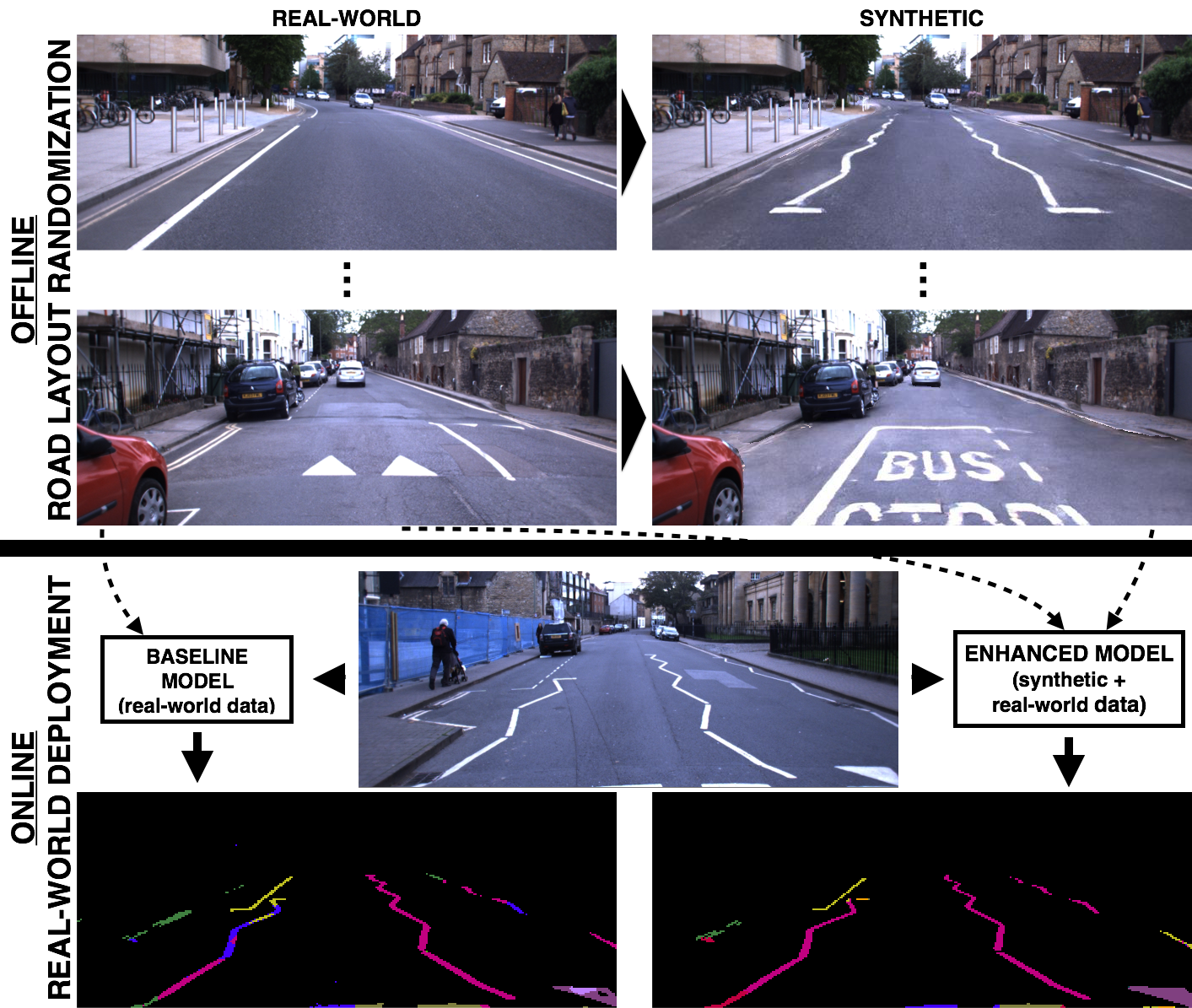}
	\caption{Road layout randomization for improved road marking segmentation, while avoiding manual labelling. \emph{Offline:} Firstly, new images for training road marking segmentation networks are automatically generated by synthesizing new road surfaces from altered semantic labels. \emph{Online:} Subsequently, mIoU of the segmentation of rare road markings (e.g. zigzags shown in pink) is improved by more than $12$ percentage points during \emph{real-world} deployment by the enhanced model trained on a hybrid dataset when compared to a baseline model only trained on the real-world dataset.}
	\label{fig:introfig}
\end{figure}

Even if more efficient hand-labelling techniques become available in the future, we still face the issue of \emph{edge cases} that appear very infrequently in regular driving.
In the context of road marking segmentation, data collection during regular driving creates extremely imbalanced datasets.
For example, zigzag markings (which indicate a pedestrian crossing, Fig. \ref{fig:introfig}) are encountered rarely, but their detection is critical for safe operation.
Resampling or applying a class-weighted loss function are not viable solutions for small, hand-labelled datasets, since these simply contain insufficient examples of rare classes for proper generalization.
Retrieving more examples is labour intensive in terms of driving and labelling time.
Consequently, trained classifiers show decreased performance on infrequently-occurring classes \cite{lee2017vpgnet}.

The latter problem could be solved by creating a virtual environment (i.e. simulator), in which the desired road markings can be reproduced as many times as necessary.
However, this introduces several new challenges.
Firstly, even though state-of-the-art simulators can appear realistic to the human-eye, their fidelity lacks the richness and complexity of the real world and consequently there is still an apparent domain gap between simulated environments and their real-world equivalent.
As a result, domain adaptation techniques need to be applied for real-world deployment \cite{chen2018learning, dundar2018domain}.
Secondly, although we might be able to generate simulated environments from real-world data in the future \cite{cura2018streetgen}, at present their design remains a manual, costly, and time-consuming task.
Besides, since urban environments can vary substantially between countries, there is a need for highly-configurable virtual worlds, which increases the labour cost.

Recently, alternative methods have been developed \cite{wang2018highresolution} to synthesize new, photo-realistic scenes for a domain of interest by employing Generative Adversarial Networks (GANs).
These approaches require relatively little human effort and can easily extend to all kinds of different conditions \cite{park2019semantic}.
This provides the ability to generate large-scale datasets for semantic scene understanding in a domain of interest at low cost.
Most of these frameworks take real-world scenes and augment them by placing or removing objects (e.g. cars, pedestrians, etc.).
This can be done randomly \cite{tremblay2018training} or more naturally by learning from real-world examples \cite{lee2018context, prakash2018structured}.

Similarly, we place instances of chosen road markings into newly-synthesized, photo-realistic scenes, which are then used to train a road marking segmentation network.
In this way, we generate sufficient examples of \emph{rare} road marking classes to achieve the generalization performance required during real-world deployment, as visualized in Fig. \ref{fig:introfig}.
However, placing new road markings coherently into the scene is difficult, since there are many dependencies such as the type of road / intersection, traffic lights, parked cars, etc. that need to be taken into account.
We avoid solving this hard problem by employing the principles of domain randomization \cite{tobin2017domain}.
More concretely, we place road markings at random places on the road surface, not necessarily coherent with other elements in the scene.
In this way, we perform \emph{road layout randomization}.
Real-world scenes encountered during deployment then appear as samples of the broadened distribution on which the model was trained.

We demonstrate quantitatively that training on these synthetic labels improves mIoU of the segmentation of rare road marking classes, for which it is expensive to attain sufficient real-world examples, during real-world deployment in complex urban environments by more than $12$ percentage points.
To take full advantage of the synthetic labels we introduce a new class-weighted cross-entropy loss which balances the training.
Furthermore, we show qualitatively that the segmentation performance for other classes is retained.

We make the following contributions in this paper:
\begin{itemize}[leftmargin=*]
	\item We present a method for generating large-scale road marking datasets for a domain of interest by leveraging principles of domain randomization, while avoiding expensive manual effort.
	\item We introduce a new class-weighted cross-entropy loss to balance the training on synthetic datasets with large class-wise imbalance in terms of their occurrence.
	\item We demonstrate a real-time framework for improving the segmentation of (rare) road marking classes in \emph{real-world}, complex urban environments.
\end{itemize}

\section{Related Work}
\subsubsection*{\textbf{Road Marking Segmentation}} Deep networks are increasingly used to perform lane detection in highway scenarios \cite{de2019end, garnett20183d, ghafoorian2019el}.
However, the urban environments and road markings targeted in this paper are substantially different and more complex, and thus require a different approach.
This problem has seen significantly fewer deep learning solutions, due to a lack of large-scale datasets containing road markings.
The first large-scale semantic road marking dataset was recently introduced in \cite{huang2018the}, however it is extremely expensive to manually expand this to all environments and conditions.

Road marking segmentation as demonstrated in \cite{lee2017vpgnet} is closest to the application of this paper.
The authors train a network for semantic road marking segmentation and improve their results by predicting the vanishing point simultaneously.
In contrast to this paper, they require thousands of hand-labelled images, which is very labour expensive.
Alternatively, the authors of \cite{hoang2019deep} hand-label road markings such as arrows and bicycle signs and train an object detection network to predict bounding boxes instead of pixel segmentations.
In previous work \cite{bruls2018mark} (includes more extensive review), we have introduced a weakly-supervised approach for binary road marking segmentation, which is used here to acquire road marking labels for real-world scenes.


\subsubsection*{\textbf{Synthetic Training for Automated Driving Tasks}}
To prevent costly and time-consuming manual labelling of training data, many approaches leverage synthetic datasets.
Early works trained on purely virtual data to perform object detection \cite{gaidon2016virtualworlds, johnsonroberson2017driving} or semantic segmentation \cite{chen2018learning, dundar2018domain}.

However, virtual data lacks the richness and complexity of the real world. A possible alternative is to augment real-world data.
For the task of semantic segmentation this means either generating new, photo-realistic images from semantic labels \cite{liu2018pixel, li2018diverse, wang2018highresolution} or enriching semantic labels with virtually-generated information \cite{geng2018part}.
Both of these principles are applied in this paper.
For object detection tasks, the main difficulty is to place the (dynamic) objects coherently into the scene.
The simplest solution is random object placement (i.e domain randomization) \cite{tremblay2018training}.
Alternatively, the authors of \cite{alhaija2018augmented, alhaija2018geometric} place photo-realistic, synthetic cars into real-world images by taking into account the geometry of the scene.
The most recent approaches \cite{khirodkar2018vadra, prakash2018structured, lee2018context, fang2018simulating} learn context-aware object placement from real-world examples.
However, placing dynamic objects, such as pedestrians, seems less complex than road markings, because the space of realistic solutions is less restrictive.
Therefore, we place road markings randomly onto the road surface in this paper.


\subsubsection*{\textbf{Scene Manipulation}}
Recently, several approaches have been introduced for more complex scene manipulation, beyond simple augmentation.
Additional sensor modalities are used in \cite{li2019aads} to offer the flexibility (e.g. different view points) of a virtual simulator, while generating data with the fidelity and richness of real-world images.
The authors of \cite{fremont2018scenic} introduce a probabilistic programming language to synthesize complex scenarios from existing domain knowledge.
Another system \cite{liu2019a} offers similar levels of control, while the camera sensor is modelled accurately at the same time.
These frameworks potentially offer a way to generate improved training data for our approach.


\section{Generating Synthetic Training Pairs}
\label{sec:synthetic}
\begin{figure}[!t]
	\centering
	\includegraphics[width=\columnwidth]{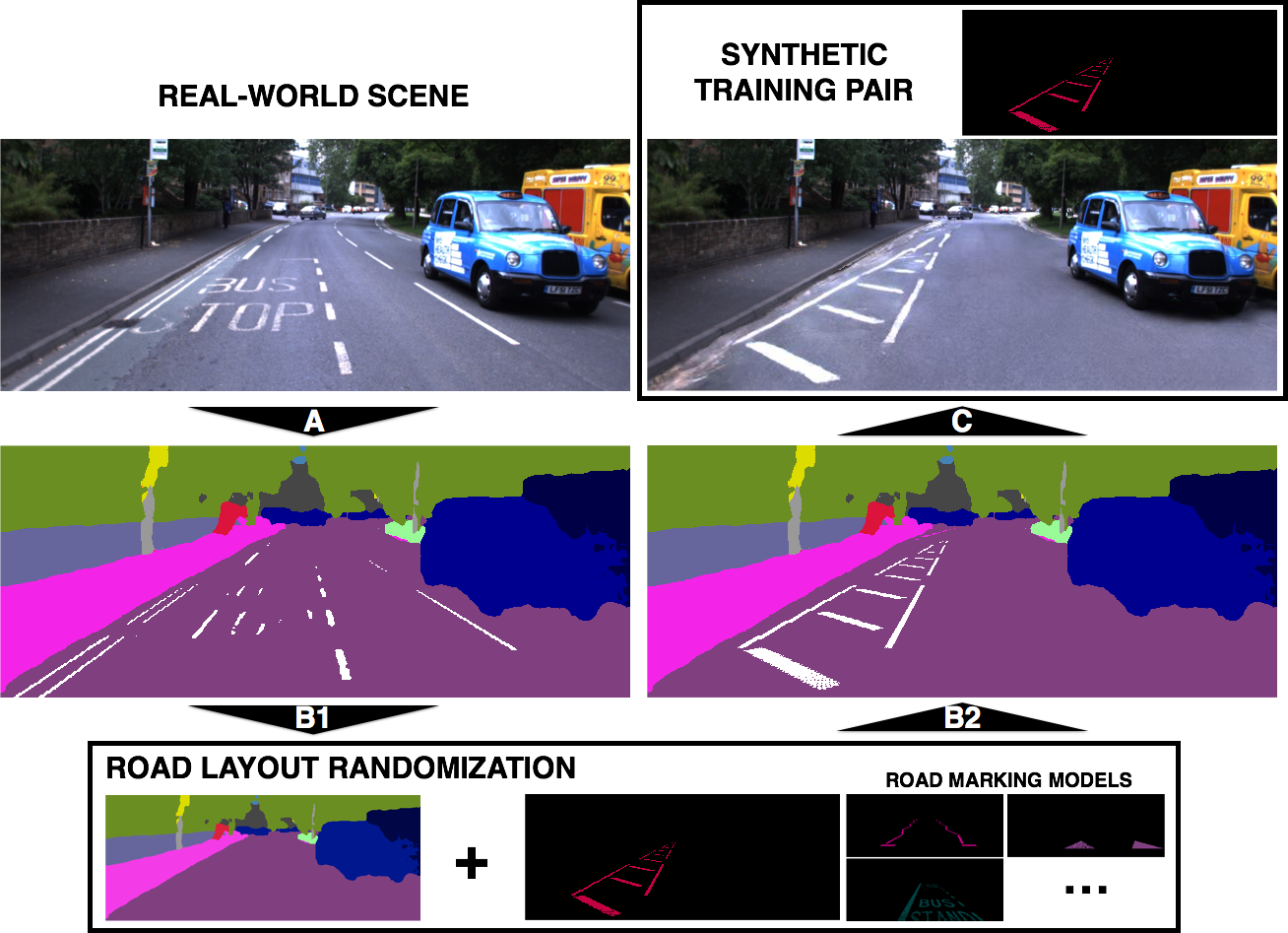}
	\caption{
	Road layout randomization: generating synthetic training data based on real-world scenes. The process has the following steps (as described in the respective subsections of Section \ref{sec:synthetic}): (A) semantic segmentation of the real-word scene is acquired, (B1) the road markings are removed and replaced with road surface, (B2) instances of chosen road markings (modelled according to the UK Highway Code) are placed randomly on the road surface, and finally
	(C) the road surface of the original image is replaced with a GAN-synthesized, photo-realistic alternative based on the altered semantic label. The composite image is then paired with the generated road marking label.}
	\label{fig:rlr}
\end{figure}

In this section, we explain in detail how to generate synthetic training pairs for road marking segmentation networks to improve performance during real-time deployment, as shown in Fig. \ref{fig:rlr}.
We demonstrate that this framework can be employed on any driving dataset even when no ground-truth semantic or road marking labels are available.

\vspace{-0.5mm}

\subsection{Retrieving Semantic Labels for Real-World Scenes}
\label{sec:realsemantics}
\begin{figure}[!t]
	\centering
	\includegraphics[width=\columnwidth]{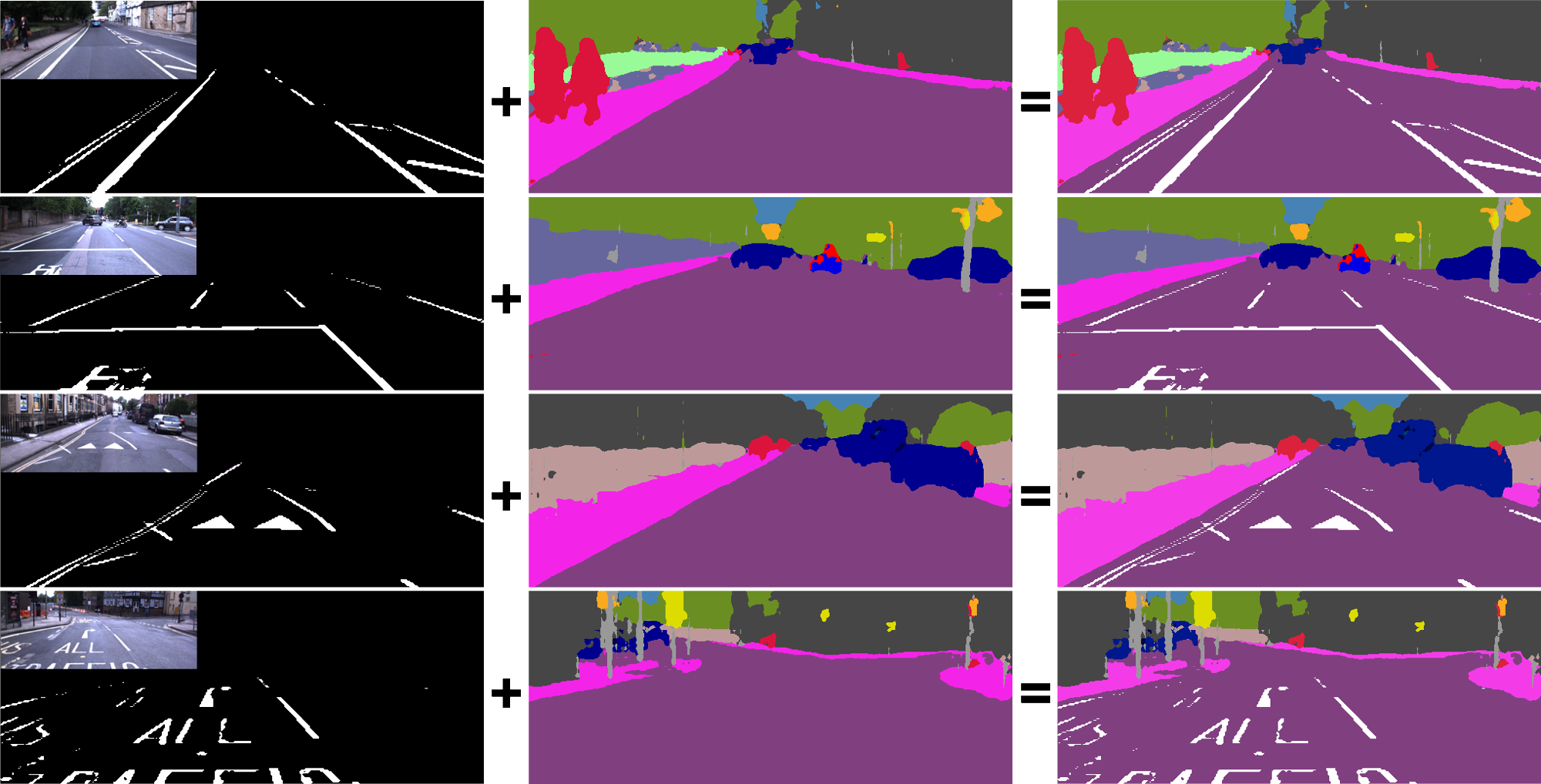}
	\caption{We augmented the Oxford Robotcar Dataset with semantic labels including road markings to train the CGAN described in Section \ref{sec:pix2pixhd}. The semantic segmentation label is retrieved from inference with a pretrained Cityscapes model and combined with automatically generated road marking annotations from \cite{bruls2018mark}. The resulting labels are not perfect ground-truth, but they are sufficient for the task and can be acquired at low cost.}
	\label{fig:semantics}
\end{figure}

In order to generate synthetic training pairs for road marking segmentation, the road layout of semantic labels of real-world scenes is altered and from these new, photo-realistic images are synthesized.
Ground-truth semantic labels are not required for the domain of interest, since semantic segmentation of reasonable (i.e. sufficient) quality can be acquired from a model pretrained on the Cityscapes dataset\footnote{\url{https://github.com/tensorflow/models/blob/master/research/deeplab/g3doc/model_zoo.md}}.
In this way, we retrieve semantic labels of real-world scenes from the Oxford RobotCar dataset \cite{maddern20171}, as shown in Fig. \ref{fig:semantics}.

Unfortunately, the available model is not trained to segment road markings (Cityscapes does not contain road marking masks).
However, semantic labels including road markings and their corresponding real-world images are necessary to train the GAN described in Section \ref{sec:pix2pixhd}.
We prevent manual labelling of road markings by employing the techniques of \cite{bruls2018mark} to generate large quantities of road marking annotations automatically.
Because these annotations are generated automatically, they are not equivalent to the ground-truth, however they have proven to be sufficient for training purposes if regularization techniques are applied.
The road markings are added to the semantic labels acquired from the Cityscapes model, as visualized in Fig. \ref{fig:semantics}.

\begin{figure*}[!t]
	\centering
	\includegraphics[width=\textwidth]{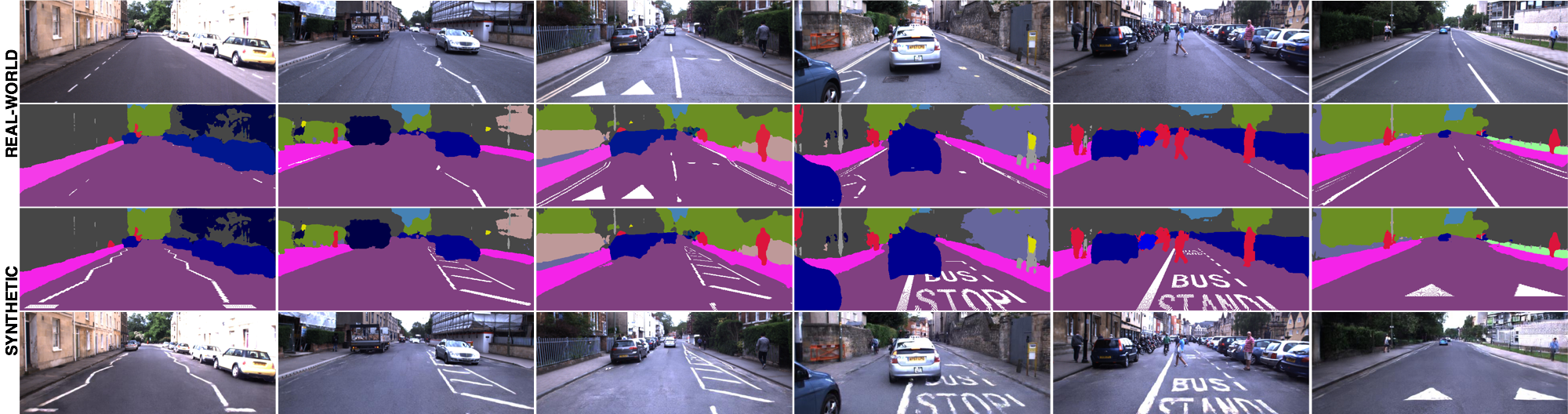}
	\caption{Examples of newly-synthesized training images for several rare road marking classes (i.e. zigzag, diagonal stripes, bus stop, and small warning triangles) by employing road layout randomization. The top two rows show images of real-world scenes of the Oxford RobotCar dataset together with the corresponding (partial) semantic labels (Section \ref{sec:realsemantics}). The third row visualizes the altered semantic labels in which instances of chosen road markings are placed randomly on the road surface (Section \ref{sec:rlr}). The last row presents the newly-synthesized road surfaces substituted into the real-world images. The GAN is able to generate road surfaces with photo-realistic textures, lighting, and even degradation as exemplified on the letters of the bus stops.}
	\label{fig:synthetic}
\end{figure*}

\subsection{Road Layout Randomization}
\label{sec:rlr}
\vspace{-1mm}
To form new road marking training pairs, we alter the road layout (i.e. road markings) of the retrieved semantic labels and subsequently synthesize a new corresponding image.
In order to rebalance datasets collected during regular driving, we create new semantic labels with road markings which occur relatively infrequently in the real world (e.g. pedestrian crossing, arrows, etc.). 
By training the road marking segmentation network on the rebalanced dataset, the goal is to improve the performance for these respective \emph{rare} classes, while at the same time retaining the overall performance.

As mentioned before, the type and placement of road markings is dependent on many factors of the scene such as the type of road, traffic lights, and even the traffic participants.
Altering all of these coherently according to the real world is difficult and seems similar in terms of complexity to designing a simulator.
Therefore, we choose to leverage domain randomization principles \cite{tremblay2018training}.
We vary position (and scale accordingly), rotation, quantity, and partial occlusion of the road markings that are placed into the environment and in that way perform \emph{road layout randomization} to create vast quantities of new training pairs automatically.
For accurate placement, we use the camera sensor calibration of the vehicle and assume that the road surface is planar and horizontal.
Training the network on many randomly-generated pairs improves generalization in newly-encountered, real-world scenes, which then appear as variations of the distribution on which the network was trained.

Concretely, we start by erasing the original road markings from the real-world semantic labels and subsequently place a new road marking instance onto the cleared road surface.
The classes are realistically modelled according to the UK Highway Code so that their shape, size, colour and configuration (e.g. zigzags appear in dual or triple configurations) resemble the real world.
Some examples for different classes of rare road markings are given in Fig. \ref{fig:synthetic}.

\subsection{Synthesizing Photo-Realistic Images}
\label{sec:pix2pixhd}
In order to create a synthetic training pair, we train a Conditional Generative Adversarial Network (CGAN), as introduced in \cite{wang2018highresolution}, to synthesize a photo-realistic RGB image for the altered semantic label (from Section \ref{sec:rlr}).
In this framework the generator $G$ aims to synthesize the RGB images, while the discriminator $D$ tries to distinguish synthesized from real-world images.
The CGAN is trained in a supervised setting using real-world images and corresponding semantic labels retrieved in Section \ref{sec:realsemantics}.
After the training is completed a photo-realistic image can be synthesized by the generator from the altered semantic labels generated in Section \ref{sec:rlr}, as shown in Fig. \ref{fig:synthetic}.

More specifically, the framework incorporates several advancements over previous works which make it possible to generate higher-resolution images.
Firstly, the generator architecture follows a traditional downsample-bottleneck-upsample model, but splits into a global generator and a local enhancer, where the local component is forced to learn high-resolution details for the stabilized features of the global component.
Secondly, to overcome discriminator capacity limitations which arise from training with high-resolution images, the framework incorporates three similar discriminators that work on different scales.
The discriminators with bigger receptive field enforce more globally consistent image generation, while the smaller receptive fields steer the generator towards more realistic, fine-level details.
Lastly, the traditional GAN loss is augmented to include a feature matching loss based on the discriminator.
Formally, following the architecture described in \cite{wang2018highresolution}, given $K=3$ discriminators $D_k$, each operating on a different scale, along with the input and label images $I^\text{SEG}_\mathrm{input}$ and $I^\text{RGB}_\mathrm{label}$, respectively, the final objective to be minimized is:
\begin{equation}
\begin{split}
   \mathcal{L}_{\mathrm{tot}} = \min_{G} ( ( \max_{D_1,D_2,D_3}\sum_{k=1,2,3}^{}\mathcal{L}_{\mathrm{GAN}}(G,D_{k}) )  + \\ \lambda_{\mathrm{FM}}\sum_{k=1,2,3}^{}\mathcal{L}_{\mathrm{FM}}(G,D_{k})+\lambda_{\mathrm{VGG}}\mathcal{L}_{\mathrm{VGG}}(G)).
\end{split}
\end{equation}
Here, $\mathcal{L}_{\mathrm{GAN}}(G,D_{k})$ represents the usual GAN loss (see \cite{wang2018highresolution}) defined over $K$ scales, $\mathcal{L}_{\mathrm{FM}}(G,D_{k})$ is the discriminator feature loss defined over $K$ scales:
\begin{equation}
 \mathcal{L}_{\mathrm{FM}(G,D_{k})}=\sum_{i=1}^{l_{D}}\frac{1}{w_{i}}{\lVert D_{k}(I^\text{RGB}_\mathrm{label})_{i} - D_{k}(G(I^\text{SEG}_\mathrm{input}))_{i} \rVert}_{1},
\end{equation}
with $l_{D}$ defining the number of layers from the discriminator used in the discriminator feature loss and $\mathcal{L}_{\mathrm{VGG}}(G)$ being the perceptual loss:
\begin{equation}
 \mathcal{L}_{\mathrm{VGG}(G)}=\sum_{i=1}^{l_{P}}\frac{1}{w_{i}}{\lVert \mathrm{VGG}(I^\text{RGB}_\mathrm{label})_{i} - \mathrm{VGG}(G(I^\text{SEG}_\mathrm{input}))_{i} \rVert}_{1},
\end{equation}
with $l_{P}$ defining the number of layers from an ImageNet-trained network (in this case VGG16) used in computing the perceptual loss. The factors $w_i=2^{l-i}$ are utilized to scale the weight of each network layer used in computing the losses.
We train the model on $3351$ overcast training pairs while using the settings as specified in \cite{wang2018highresolution} to generate images with a resolution of $256\times640$.

Unfortunately, the RobotCar dataset does not contain any boundary or instance labels (as used in \cite{wang2018highresolution}) necessary to generate sharp, high-quality images.
Consequently, the generated images can be smudgy around object boundaries (e.g. rows of parked cars are merged because of the image perspective, as exemplified in \cite{wang2018highresolution}) and contain unnatural artifacts.
Therefore, we choose to substitute only the newly-generated road surface and keep the rest of the original image intact.
The RobotCar dataset contains sufficient real-world images so that no background duplicates have to exist in the new road marking dataset.
In this way, we are able to generate a large-scale urban datasets for road marking segmentation, while avoiding expensive manual labelling.

The above-described framework can easily be extended to different (weather and lighting) conditions by training condition-specific models.
If it is not possible to retrieve semantic labels of sufficient quality under difficult conditions, a state-of-the-art invertible generator, that can transform the images into the desired appearance similar to \cite{porav2018adversarial, park2019semantic}, can be employed.
In this way the semantic label acquired from the overcast image can be paired with an image which resembles a different weather or lighting condition.

\section{Training for Road Marking segmentation}
\label{sec:segmentation}
In this section, the network trained for road marking segmentation is described in detail, along with some important considerations that have to be taken into account when rebalancing datasets.

\subsection{Network Architecture}
Deep networks for road marking segmentation have several advantages over traditional heuristic or shallow-learning pipelines.
Firstly, they are more robust to spatial deformations, degradation, and partial occlusion.
Secondly, the scene context can be leveraged to improve semantic segmentation and thereby understand the road rules.
For instance, similarly-shaped road markings (e.g. lane separators and separators that mark a parking spot) can be classified differently based on their place in the scene and relationship with other objects, whereas this is difficult to accomplish with traditional rule-based systems.

We train a U-Net model \cite{ronneberger2015unet}, but include batch normalization and dropout as regularization techniques.
These are paramount in our framework, since we train on partial labels that are generated automatically.
Dropout allows the network to extend its prediction towards road marking pixels that were wrongly assigned to the background in the partial labels, because they share more similarities with the road marking class than the background class.
The architecture and training settings used are similar to our previous work \cite{bruls2018mark}, with the major exception that the output now predicts multiple classes of road markings instead of a binary segmentation.
More specifically, the output of the network is computed by applying a channel-wise softmax activation over the final feature maps and assigning a class to each respective pixel by taking the channel-wise $\argmax$ over the output channels, yielding a one-channel discrete class activation map.

At run time, the Tensorflow implementation of the network performs inference on an input image in real-time (${\sim}62.5$\,Hz) on an NVIDIA TITAN Xp GPU.

\subsection{Balancing of the Classes}
\label{balancing}
As mentioned before, datasets collected during regular driving are extremely imbalanced in terms of the occurrences of particular road marking classes.
For instance, zigzag markings are only found in ${\sim}7\%$ of the images, whereas lane separators occur in ${\sim}70\%$.
Solutions such as resampling the dataset or applying a class-weighted loss function are not viable for small, hand-labelled datasets, because they simply contain an insufficient number of examples of the rare classes to generalize well to unseen cases during deployment.

In this paper, we opt for a different approach in which we synthesize new training pairs for rare classes automatically and add them to an existing dataset.
This ensures that there are enough examples of these classes for the network to learn from.
However, it is not obvious how to produce a rebalanced dataset including synthetic training pairs that is optimal for training.
To counteract the fact that we might add too many synthesized training pairs, we experiment with three types of class-weighted cross-entropy losses:
\begin{enumerate}[leftmargin=*]
    \item Equal weighting (EQ) of all classes irrespective of their occurrence in the dataset.
    \item Median frequency balancing (FB) \cite{eigen2015predicting}, in which each pixel is weighted by
    \begin{equation}
        w_c = \dfrac{\mathrm{median}\left(F\right)}{f_c}, \\
    \end{equation}
    where $F = \{f_1, \dots , f_C\}$ with $f_c$ denoting the total number of pixels of class $c$ divided by the total number of pixels in labels where $c$ is present and $C$ the total number of classes.
    \item Median total balancing (TB), in which each pixel is weighted by
    \begin{equation}
        w_c = \dfrac{\mathrm{median}\left(G\right)}{f_c + n_c},
    \end{equation}
    where $G = \{f_1 + n_1, \dots, f_C + n_C\}$ with $f_c$ equivalent to 2) and $n_c$ denoting the number of labels in which class $c$ is present divided by the total number of training pairs.
\end{enumerate}
It is important to note that median frequency balancing only corrects for the fact that some classes naturally occupy less pixels in the images.
For instance, dotted lines indicating a pedestrian crossing are smaller in accumulated area than an alternative zebra crossing.
However, median frequency balancing does not account for imbalance in occurrences across the dataset; whether ${\sim}7\%$ of the images contain zigzag markings or ${\sim}70\%$, the weight remains the same as long as their pixel size remains equivalent.
This is not ideal, since we artificially create an imbalance in the number of occurrences by adding labels of specific classes.
The third weighting function, introduced in this paper, is designed to take this into account, balancing the average pixel area as well as the imbalance in occurrences across the dataset.

\section{Experimental Results}
\label{sec:experiments}

In this section we describe the experimental setup and the datasets that we have created, before we present the quantitative and qualitative results.

\subsection{Experimental Setup}
\begin{figure*}[!t]
	\centering
	\includegraphics[width=\textwidth]{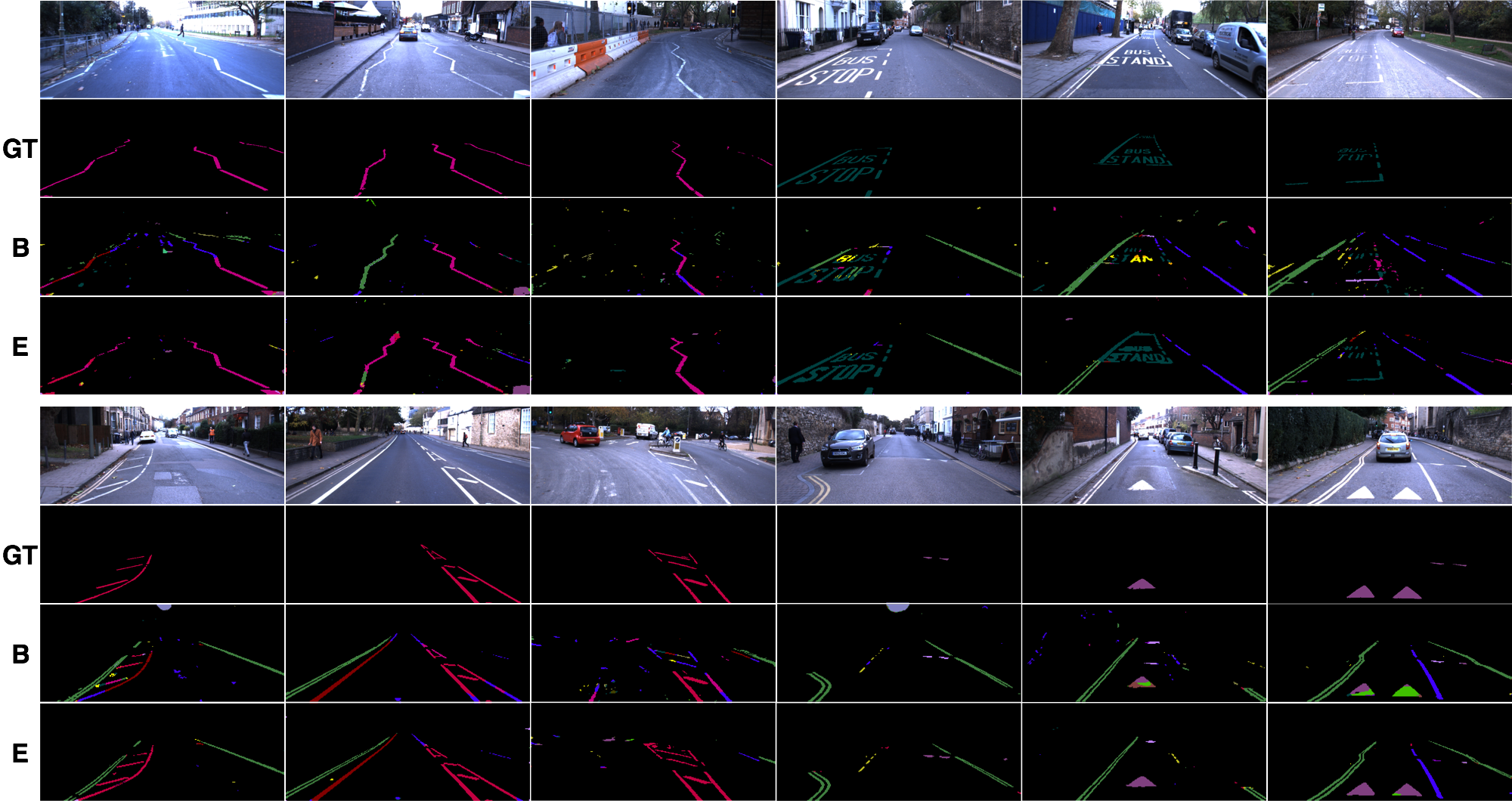}
	\caption{Road marking segmentation (full set of classes) in traffic environments with rare classes. The \emph{top two} rows of each scene show the input image together with the corresponding ground-truth (GT) label of the rare class, which is used for quantitative evaluation. The \emph{bottom two} rows of each scene depict the segmentation results for the best performing baseline (B) and enhanced model (E), respectively. The enhanced model provides more consistent and correct segmentation of the rare classes, while retaining reasonable and sometimes achieving improved performance for other classes (e.g. \emph{green} double boundaries, \emph{blue} separators, \emph{yellow} parking spot separators, etc.).}
	\label{fig:qualitative}
\end{figure*}

We have selected four types of rare road markings for evaluation: bus stops, diagonal stripes (must not enter), small warning triangles, and zigzag markings.
These classes function as a proof of concept, but the framework can be applied to any class (i.e. model) of road markings.
For quantitative pixel-wise evaluation, we have hand-labelled $102$, $102$, $96$, and $102$ \emph{real-world} images containing bus stops, diagonal structures, small warning triangles, and zigzag markings, respectively.
Note that in these images only these respective classes were labelled and all other classes present were ignored (see Fig. \ref{fig:qualitative}).
While we train all models to predict the \emph{full} set of $20$ different road markings and show these results qualitatively, we only evaluate the four selected classes quantitatively.
We define the pixel-wise metrics $\mathrm{PRE=\frac{TP}{TP+FP}}$, $\mathrm{REC=\frac{TP}{TP+FN}}$, $\mathrm{F_1=2*\frac{PRE*REC}{PRE+REC}}$, and $\mathrm{IoU = \frac{TP}{TP+FP+FN}}$ with $\mathrm{TP}$, $\mathrm{FP}$, and $\mathrm{FN}$ denoting the true positive, false positive, and false negative pixels, respectively.
In contrast to binary classification, all metrics are evaluated at the operating point defined by taking the channel-wise $\argmax$ over the multi-class output on a per image basis and averaged over the test set, without any further fine tuning of the operating characteristics.
Furthermore, we have hand-labelled $25$ real-world images for each respective class for validation.
We train until convergence and select the epoch for testing in which the mIoU is highest among the evaluations on the validation set.
It should be noted that road marking segmentation is arguably a harder task than scene segmentation, because road marking elements are fairly small in general, often degraded, and the different types share many visual and geometric similarities.
State-of-the-art approaches achieve a mIoU of around $40\%$, however a benchmark has only been established recently \cite{huang2018the}.

\begin{figure}[!t]
	\centering
	\includegraphics[width=\columnwidth]{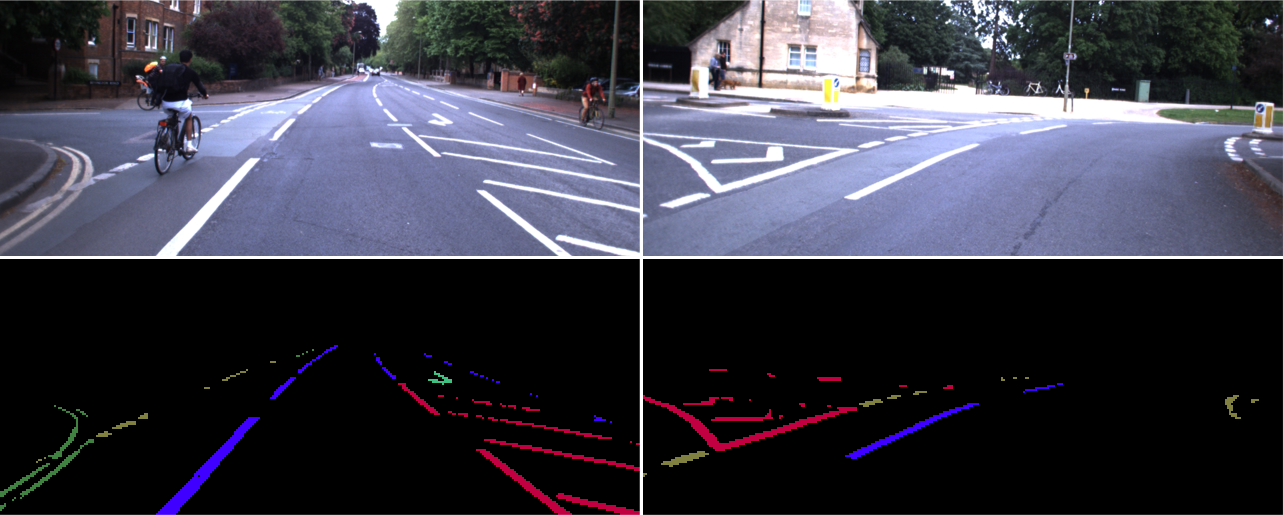}
	\caption{Examples of the partial labels created by semantically classifying the binary annotations of \cite{bruls2018mark}. Although not perfect ground-truth, these labels can be used to train a baseline model to predict the full set of road markings.}
	\label{fig:partial}
\end{figure}

As a reasonable baseline, $1000$ partial, binary labels generated by \cite{bruls2018mark} collected during regular driving were hand-labelled class-wise.
Although not equivalent to the ground-truth, we have proven in \cite{bruls2018mark} and will demonstrate again in Section \ref{sec:quali} that these labels are sufficient to achieve full segmentation, when regularization techniques are applied.
A few examples are given in Fig. \ref{fig:partial}.
The labels contain the $20$ different types of road markings, so that the network functions as a full road marking segmentation system.
However, many classes occur too infrequently to achieve state-of-the-art performance, because the network fails to generalize to new scenarios during deployment.
For instance, the baseline dataset only contains $63$, $109$, $39$, and $74$ images with bus stops, diagonal stripes, small warning triangles, and zigzag markings, respectively.
For the other experiments, we add synthetic training pairs of the four classes to the baseline dataset.
In this way, the network still predicts all 20 classes, but is given a sufficient number of labels of the rare classes to improve generalization during real-world deployment.

\begin{figure}[!t]
	\centering
	\includegraphics[width=\columnwidth]{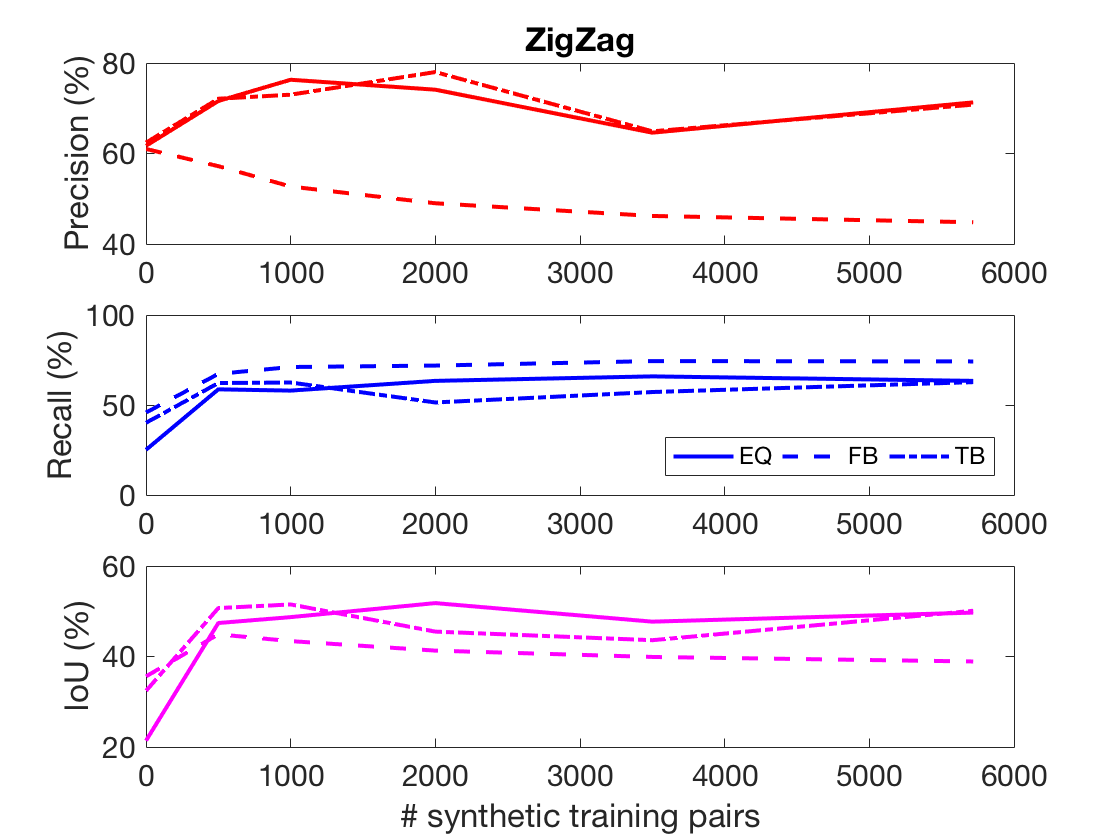}
	\caption{Pixel-wise performance of zigzag segmentation when training with different cross-entropy losses for a variable number of synthetic images added to the baseline dataset.}
	\label{fig:graph}
\end{figure}

\begin{table*}
    \caption{Pixel-wise Performance for Rare Classes for the Baseline (B) and Enhanced (E) Models}
    \begin{adjustbox}{width=\textwidth,center}
    \begin{tabular}{@{}c c cccc cccc cccc cccc cccc@{}}\toprule
    & & \multicolumn{4}{c}{\textbf{BUS STOP}} & \multicolumn{4}{c}{\textbf{DIAGONAL}} & \multicolumn{4}{c}{\textbf{TRIANGLE}} & \multicolumn{4}{c}{\textbf{ZIGZAG}} & \multicolumn{4}{c}{\textbf{MEAN}}\\
    \cmidrule(lr){3-6}
    \cmidrule(lr){7-10}
    \cmidrule(lr){11-14}
    \cmidrule(lr){15-18}
    \cmidrule(lr){19-22}
    \textbf{Model} &\textbf{Loss} &\textbf{PRE} &\textbf{REC} &$\mathbf{F_1}$ &\textbf{IoU} &\textbf{PRE} &\textbf{REC} &$\mathbf{F_1}$ &\textbf{IoU} &\textbf{PRE} &\textbf{REC} &$\mathbf{F_1}$ &\textbf{IoU} &\textbf{PRE} &\textbf{REC} &$\mathbf{F_1}$ &\textbf{IoU} &\textbf{PRE} &\textbf{REC} &$\mathbf{F_1}$ &\textbf{mIoU}\\ \midrule
    B &EQ   &61.6 &17.8 &27.6 &16.1                             &59.1 &24.6 &34.7 &21.8                    &60.7 &41.1 &49.0 &34.4                             &65.9 &22.5 &33.6 &20.1                    &61.8 &26.5 &36.2 &23.1\\
    B &FB   &64.7 &26.3 &37.3 &22.8                             &58.8 &31.0 &40.6 &26.4                    &60.1 &47.7 &53.2 &36.9                             &64.7 &34.8 &45.3 &29.5                    &62.1 &35.0 &44.1 &28.9 \\
    B &TB   &62.2 &19.1 &29.2 &17.1                             &58.4 &33.3 &42.4 &27.7                    &59.9 &51.6 &53.4 &39.4                             &62.3 &31.3 &41.7 &26.5                    &60.7 &33.8 &42.2 &27.7 \\[5pt]
    E &EQ   &\textbf{74.8} &28.5 &41.2 &26.3                    &\textbf{73.0} &40.9 &52.4 &35.8           &\textbf{61.4} &46.9 &53.2 &38.4                    &\textbf{69.4} &49.8 &58.0 &40.7           &\textbf{69.7} &41.5 &51.2 &35.3 \\
    E &FB   &54.8 &\textbf{62.9} &\textbf{58.6} &\textbf{40.1}  &45.8 &\textbf{67.4} &54.6 &39.1           &46.9 &73.8 &57.3 &37.9                             &51.0 &\textbf{66.6} &57.8 &40.3           &49.6 &\textbf{67.7} &57.1 &39.4 \\
    E &TB   &58.5 &55.3 &56.8 &39.2                             &51.4 &59.1 &\textbf{55.0} &\textbf{40.2}  &50.8 &\textbf{75.1} &\textbf{60.6} &\textbf{43.4}  &61.4 &57.3 &\textbf{59.3} &\textbf{42.5}  &55.5 &61.7 &\textbf{57.9} &\textbf{41.3} \\
    \bottomrule
    \end{tabular}
    \end{adjustbox}
    \label{tab:quant}
\end{table*}

\subsection{Quantitative Evaluation}
\label{sec:quanti}
In order to understand how the number of added synthetic images influences the performance, we have added different numbers of synthetic zigzag pairs to the baseline dataset, while keeping the other classes constant.
The results for the three different cross-entropy losses are presented in Fig. \ref{fig:graph}.
The following key observations can be made:
\begin{itemize}[leftmargin=*]
    \item Adding as little as 500 synthetic training pairs already makes a substantial difference in terms of overall performance.
    \item Adding more than 2000 synthetic training pairs does not provide extra benefits in general. Further performance increase beyond this level might require higher-quality, more diverse, more coherent synthetic images.
    \item FB struggles to balance training as more synthetic pairs are added, due to the fact that it does not account for occurrence imbalance across the dataset. The precision drops significantly as the network learns from an abundance of zigzag markings and starts classifying other classes incorrectly as zigzag.
    \item TB alleviates the precision drop of FB, but does not outperform EQ consistently among all metrics.
\end{itemize}
Assuming that these observations hold similarly for the other classes, 1000 synthetic training pairs of each respective class were added to the baseline dataset as a proof of concept to train enhanced networks (with the different loss functions).
From the results, as presented in Table \ref{tab:quant}, the following key observations can be made:
\begin{itemize}[leftmargin=*]
    \item By adding synthetic training pairs, IoU performance similar to the state-of-the-art can be achieved when only very few real-world examples are available. mIoU is increased by $12.4\%$ (comparing the best baseline and enhanced models) without using any manual labelling effort.
    \item The enhanced networks always achieve better overall performance (i.e IoU) by a substantial margin for the equivalent cost function. Segmentation performance can thus be boosted cheaply by the presented framework.
    \item TB outperforms FB in terms of $\text{F}_1$ and IoU in general, because it accounts for the class imbalance across the dataset that was artificially created by adding synthetic pairs. TB offers a good trade-off between high precision achieved by EQ and high recall achieved by FB.
\end{itemize}


\subsection{Qualitative Evaluation}
\label{sec:quali}
\vspace{-1mm}
In Fig. \ref{fig:qualitative}, the best baseline and enhanced models are compared qualitatively for different traffic scenes.
All networks are trained to predict the full set of $20$ different road marking classes, however scenes with the respective rare classes are selected for visualization.

It is clear that adding synthetic images to the training set results in more consistent and correct segmentation of the rare classes, while retaining reasonable and sometimes achieving improved performance for other classes.
The latter could be caused by the general increase of the number of training examples and/or better balancing of the cost function.
The enhanced model trained with TB offers more satisfying (i.e. less noisy) visual results than the baseline model trained with FB.
Furthermore, it is clear that full segmentation of the road marking elements is possible from partial labels when regularization techniques are applied correctly.
Thus, this framework offers an effective and efficient step towards a road marking classification system for automated driving pipelines.


\section{Conclusion}
We have presented a weakly-supervised approach for improving road marking segmentation in complex urban environments.
To this end, we alter semantic labels of real-world scenes with instances of chosen road markings using domain randomization principles and synthesized corresponding, photo-realistic images to generate vast quantities of synthetic training pairs, thereby avoiding the need for expensive manual labelling.
During deployment, we predict $20$ classes of road markings in real time and we have demonstrated quantitatively that this framework improves mIoU of rare classes by more than $12$ percentage points and thus reaches state-of-the-art performance with very few real-world labels.
This is achieved by introducing a new class-weighted cross-entropy loss to balance the training of synthetic datasets.
Furthermore, we have shown qualitatively that the segmentation performance for other classes is retained.
The presented framework can easily be extended to include other classes or work under different conditions and results can be expected to improve as more advanced synthesizing networks will emerge in the future.
Hence, road layout randomization is an effective and efficient technique to enhance road marking classification systems in automated driving pipelines.
\vspace{-1mm}

\section*{Acknowledgment}
The work has been supported by the EPSRC/UK Research and Innovation Programme Grant EP/M019918/1 (Mobile Autonomy: Enabling a Pervasive Technology of the Future).
We acknowledge the support of NVIDIA Corporation with the donation of Titan Xp and Titan V GPUs.


\end{document}